\documentclass[11pt]{article}

\usepackage[final]{acl}

\usepackage{amsmath}

\usepackage{times}
\usepackage{latexsym}

\usepackage{float}
\usepackage{hyperref}
\usepackage[T1]{fontenc}

\usepackage[utf8]{inputenc}
\usepackage{booktabs}
\usepackage{multirow}

\usepackage{microtype}

\usepackage{inconsolata}

\usepackage{graphicx}

\title{Aaron at SemEval-2026 Task 9: Multilingual Polarization Detection Using Transformer-Based Models with Class Weighting and Threshold Tuning}

\author{Aaron Bundi Anampiu \\
African Institute for Mathematical Sciences, South Africa\\
  \texttt{aaronbundi@aims.ac.za} }
  
\begin{document}
\maketitle
\begin{abstract}
This paper describes our submission to SemEval-2026 Task 9 on detecting multilingual, multicultural, and multievent online polarization. We address all three subtasks: binary polarization detection, polarization type classification, and manifestation identification for English and Swahili. Our approach leverages transformer-based models (RoBERTa-base for English, AfroXLMR-base for Swahili) with class-weighted loss functions to address severe label imbalance and per-label threshold tuning to optimize multi-label classification. On the test set, we achieve F1 macro scores of 0.7901 (English) and 0.7910 (Swahili) for Subtask 1, 0.4615 (English) and 0.4808 (Swahili) for Subtask 2 and 0.4791 (English) and 0.5830 (Swahili) for Subtask 3, which give competitive performance on the leaderboard, demonstrating the effectiveness of our methods for handling imbalanced multi-label polarization detection. Our error analysis reveals that models struggle with dehumanization detection and lack of empathy.

\end{abstract}

\section{Introduction}

Social media platforms have become central spaces for public discourse,
enabling millions of users to share opinions, engage in debates, and form
communities.
However, these platforms have also witnessed a concerning rise in polarized
content and messages that divide audiences along ideological, political, racial,
religious, or other lines \cite{grover2019polarization}.

Polarization poses significant societal challenges. It amplifies echo
chambers, reduces constructive dialogue, can incite real-world conflicts, and
disproportionately affects marginalized communities \cite{conover2011political}.
Automatic detection of polarizing content is therefore crucial for content
moderation, understanding radicalization mechanisms, and promoting healthier
online discourse. 

SemEval-2026 Task~9 \cite{naseem-etal-2026-polar} addresses this challenge
through three subtasks: (1)~binary detection of polarization, (2)~multi-label
classification of polarization types (political, racial/ethnic, religious,
gender/sexual, other), and (3)~multi-label identification of polarization
manifestations (stereotype, vilification, dehumanization, extreme language,
lack of empathy, invalidation). The task covers 22~languages from diverse
platforms including news websites, Reddit, blogs, and regional forums.

We focus on English and Swahili, employing a transformer-based approach with three key contributions: (1) language-specific model selection (RoBERTa ~\cite{liu2019roberta} for English and AfroXLMR base ~\cite{alabi2022adapting} for Swahili), (2) class-weighted loss functions to address severe label imbalance in multi-label tasks, and (3) per-label threshold tuning to optimize classification boundaries.  Our system achieves strong performance on all the Subtasks. It was ranked 2nd for Swahili on Subtask~3 among the 16 participating teams.

\section{Background and Related Work}

\subsection{Task Description}

SemEval-2026 Task~9 provides 3,000 to 5,000 annotated instances per language
\cite{naseem2026polarbenchmarkmultilingualmulticultural} spanning elections,
conflicts, gender rights, and migration.

\paragraph{Subtask~1} (Binary Detection) classifies text as polarized or
non-polarized based on context and overall meaning.

\paragraph{Subtask~2} (Type Classification) identifies five polarization
types: political/ideological, racial/ethnic, religious, gender/sexual, and
other (multi-label).

\paragraph{Subtask~3} (Manifestation Identification) classifies six
manifestations: stereotype, vilification, dehumanization, extreme language,
lack of empathy, and invalidation (multi-label).

\subsection{Related Work}

Prior work on detecting harmful online content has primarily focused on 
hate speech~\cite{fortuna2018survey}, toxic 
comments~\cite{wulczyn2017ex}, and offensive language~\cite{zampieri2019predicting}. 
These tasks share similarities with polarization detection but differ in 
important ways.
Early approaches used traditional machine 
learning with hand-crafted features~\cite{davidson2017automated}, including 
n-grams, sentiment lexicons, and linguistic patterns. More recent work 
has adopted deep learning, with \citet{badjatiya2017deep} demonstrating 
that LSTMs with word embeddings significantly outperform traditional 
methods.  \citet{devlin2019bert} showed that BERT-based models achieve 
better results on hate speech benchmarks.

The development of multilingual pre-trained 
models has enabled effective transfer across languages. mBERT~\cite{devlin2019bert} and XLM-RoBERTa~\cite{conneau2020unsupervised} 
demonstrated strong zero-shot cross-lingual transfer capabilities.  However, 
these models are trained primarily on high-resource languages and may 
underperform on African languages.

Recent work has developed language models 
specifically for African languages.  AfriBERTa~\cite{ogueji2021small} and 
AfroXLMR~\cite{alabi2022adapting} were pre-trained on diverse African 
language corpora, including Swahili, and have shown superior performance 
on African NLP tasks compared to general multilingual models.  We adopt 
AfroXLMR for Swahili based on these findings.

\section{System Overview}

\begin{figure*}[t]
\centering
\includegraphics[width=0.85\textwidth]{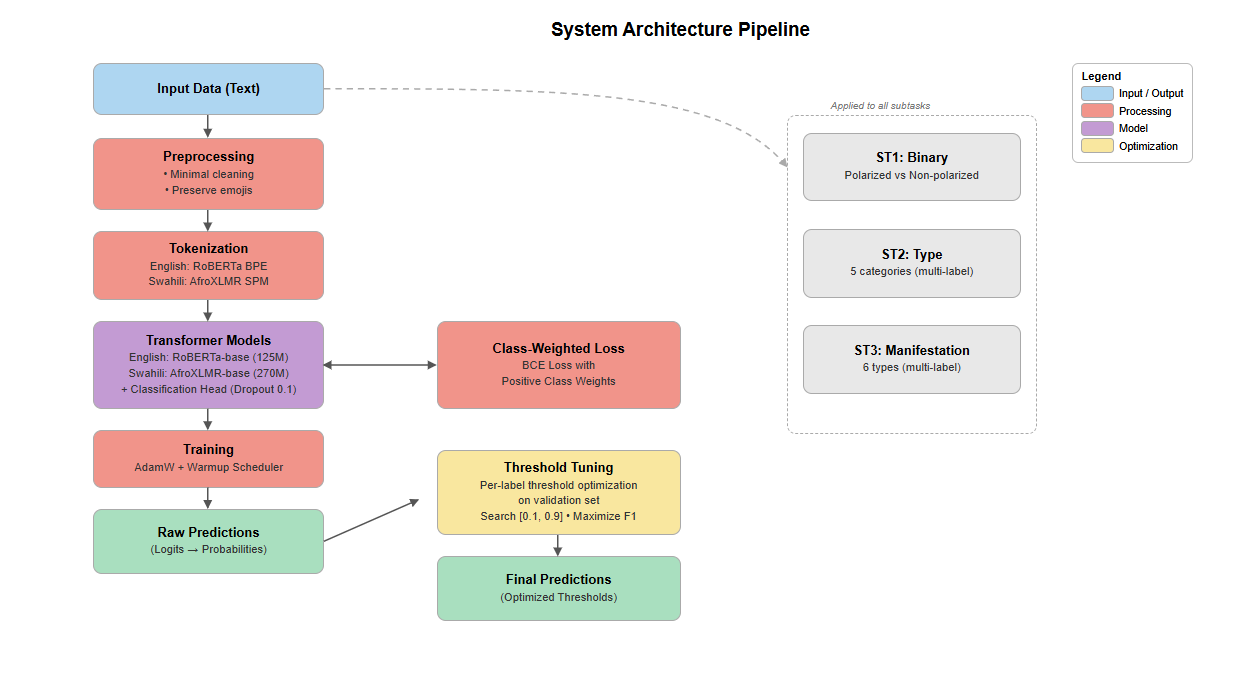}
\caption{Complete system architecture pipeline showing the four main components: preprocessing, model architecture (language-specific transformers), training with class-weighted loss, and post-training threshold tuning. The pipeline is applied to all three subtasks.}
\label{fig:architecture}
\end{figure*}

Our pipeline has four components: preprocessing, model architecture, training
with class-weighted loss, and post-training threshold tuning.
Figure~\ref{fig:architecture} illustrates the full pipeline.

\subsection{Data Preprocessing}

We perform minimal preprocessing: retain original casing, preserve emojis and
special characters that may signal polarization, and remove only control
characters and null bytes. Data are split 80/20 into train/validation using
stratified sampling to maintain class distribution.

\subsection{Model Architecture}

\paragraph{English Models:} RoBERTa-base \cite{liu2019roberta}
(125M~parameters) was selected over BERT and XLM-RoBERTa based on ablation
results (Table~\ref{tab:ablation_models}): its dynamic masking and removal of
next-sentence prediction yield the best English-specific performance, while
XLM-RoBERTa's multilingual pre-training dilutes English-specific
representations.

\paragraph{Swahili Models:} AfroXLMR-base \cite{alabi2022adapting} (270M parameters) is specifically adapted for African languages through continued pre-training on diverse African language corpora. It significantly outperforms multilingual alternatives like mBERT and XLM-RoBERTa on Swahili tasks \cite{adelani2022masakhaner}. 

\paragraph{Tokenization:}RoBERTa's byte-level BPE tokenizer (50K vocab) for
English; AfroXLMR's SentencePiece tokenizer (250K vocab) for Swahili. Maximum
sequence length is 128 tokens for Subtask~1 and 256 for Subtasks~2--3.

\paragraph{Classification head:} A dropout layer (rate~0.1) followed by a
linear layer maps 768-dimensional embeddings to 2, 5, or 6 output logits for
Subtasks 1, 2, and 3 respectively.

\subsection{Class-Weighted Loss Function}

To address severe label imbalance, we implement class-weighted binary cross-entropy loss with positive class weight for label computed as:

\begin{equation}
w_j = \frac{n_{\text{neg}}^j}{n_{\text{pos}}^j}
\end{equation}

where $n_{\text{neg}}^j$ and $n_{\text{pos}}^j$ are negative and positive sample counts for label $j$. This increases the penalty for misclassifying minority classes. For Subtask 2 English, weights range from 1.73 (political) to 41.95 (gender/sexual), reflecting the extreme rarity of gender-related polarization.

Weights for Swahili Subtask~2 reach 48.05 (gender/sexual). Training loss curves showed no instability; gradient clipping (max-norm 1.0) was applied throughout as a precaution.

\subsection{Threshold Tuning}

After training we perform per-label threshold search on the validation set,
scanning $[0.1, 0.9]$ in steps of 0.05 to maximise macro~F1 per label. This
improves macro~F1 by 0.87--5.98 points over the standard 0.5 threshold.

\subsection{Training Configuration}

We use AdamW \cite{loshchilov2019decoupled} (learning rate 2e-5, weight decay
0.01) with a linear warmup schedule, chosen as the standard optimizer for
fine-tuning transformers. Batch size is 16 for Subtask~1 and 8 with gradient
accumulation ($\times$2) for Subtasks~2--3 to fit 256-token sequences in 15~GB
GPU memory. We train for 5--8 epochs with early stopping (patience~2) based
on validation macro~F1, using FP16 mixed precision on a single GPU via Google
Colab. Random seed is fixed to~42 for all experiments.

\section{Experimental Setup}

\subsection{Data Splits}

We use provided train/dev splits. Training data are further split 80/20 using
stratified sampling (binary labels for Subtask~1; label-count strata for
Subtasks~2--3). The test set is used exclusively for final evaluation.

\subsection{Evaluation Metrics}

All subtasks use macro~F1 as the primary metric. We compare our system against
two baselines:
(i)~a \textbf{POLAR Baseline}, and (ii)~an \textbf{mBERT fine-tuned baseline} applied
uniformly across all languages without class weighting.
Table~\ref{tab:baselines} summarises this comparison.

\begin{table}[H]
\centering
\small
\begin{tabular}{@{}llccc@{}}
\toprule
\textbf{ST} & \textbf{Lang} & \textbf{POLAR Baseline} & \textbf{mBERT} & \textbf{Ours} \\
\midrule
\multirow{2}{*}{1} & EN & 0.7802 & 0.7210 & \textbf{0.7901} \\
                   & SW & 0.7571 & 0.6981 & \textbf{0.7910} \\
\midrule
\multirow{2}{*}{2} & EN & 0.3333 & 0.3810 & \textbf{0.4615} \\
                   & SW & 0.4417 & 0.3641 & \textbf{0.4808} \\
\midrule
\multirow{2}{*}{3} & EN & 0.4100 & 0.4121 & \textbf{0.4791} \\
                   & SW & 0.2205 & 0.4380 & \textbf{0.5830} \\
\bottomrule
\end{tabular}
\caption{Test macro~F1 comparison with POLAR baselines. ST = Subtask,
EN = English, SW = Swahili. Our system consistently outperforms both baselines
across all subtasks and languages.}
\label{tab:baselines}
\end{table}

\subsection{Implementation}

We implement our system using Hugging Face Transformers~4.30.0, PyTorch 2.0, and scikit-learn~1.3.0.

\section{Results}

\subsection{Main Results}

Table~\ref{tab:main_results} presents results across all Subtasks.

\begin{table}[ht]
\centering
\small
\begin{tabular}{@{}llcc@{}}
\toprule
\textbf{Subtask} & \textbf{Language} & \textbf{Val F1} & \textbf{Test F1} \\
\midrule
\multirow{2}{*}{1. Binary Detection} 
& English & 0.8017 & \textbf{0.7901} \\
& Swahili & 0.7766 & \textbf{0.7910} \\
\midrule
\multirow{2}{*}{2. Type Classification} 
& English & 0.4721 & \textbf{0.4615} \\
& Swahili & 0.4598 & \textbf{0.4808} \\
\midrule
\multirow{2}{*}{3. Manifestation ID} 
& English & 0.5226 & \textbf{0.4791} \\
& Swahili & 0.5742 & \textbf{0.5830} \\
\bottomrule
\end{tabular}
\caption{F1 macro scores on validation and test sets. Bold indicates official test scores used for ranking in codabench.}
\label{tab:main_results}
\end{table}

\paragraph{Subtask~1} achieves our strongest performance (0.7901 EN,
0.7910 SW). Swahili slightly outperforms English, likely due to a more
balanced class distribution (50.1\% polarized in Swahili vs.\ 36.5\% in
English). We ranked 11 out of 34 teams for Swahili and 24 out of 44 teams for English on this Subtask.

\paragraph{Subtask~2} presents challenges from extreme imbalance and
multi-label complexity. We achieved 0.4615 (EN) and 0.4808 (SW); the ranking for Swahili was 10
out of 22 teams and for English, 17 out of 29 teams. Swahili benefited from a more balanced label
distribution and AfroXLMR's stronger multilingual capabilities.

\paragraph{Subtask~3} yields our most competitive results: 0.4791 (EN) and
0.5830 (SW). We ranked 2nd out of 16 teams for Swahili and, for English, 10 out of 18 teams on this Subtask.

The validation-test gap ranges from 2.29 to 10.58 points. The largest
discrepancy is Subtask~3 English (4.35 points). We attribute this to two
compounding factors: (i)~\textit{threshold overfitting}---per-label thresholds
are optimised on a single 20\% validation split, making them sensitive to
its specific label distribution; and (ii)~\textit{domain shift}---the test
set may contain text from platforms or events not well-represented in
training. Cross-validation for threshold selection is a natural mitigation,
though it substantially increases compute time.

\subsection{Per-Label Analysis}

Tables~\ref{tab:subtask2_labels} and~\ref{tab:subtask3_labels} show per-label
F1 scores.

\begin{table}[ht]
\centering
\small
\begin{tabular}{@{}lcc@{}}
\toprule
\textbf{Label} & \textbf{English} & \textbf{Swahili} \\
\midrule
Political & \textbf{0.6682} & 0.3256 \\
Racial/Ethnic & 0.5082 & \textbf{0.7972} \\
Religious & 0.5957 & 0.6387 \\
Gender/Sexual & 0.3200 & 0.2590 \\
Other & 0.3000 & 0.2784 \\
\midrule
\textbf{Macro Average} & \textbf{0.4784} & \textbf{0.4598} \\
\bottomrule
\end{tabular}
\caption{Validation F1 per label, Subtask~2.}
\label{tab:subtask2_labels}
\end{table}

\begin{table}[ht]
\centering
\small
\begin{tabular}{@{}lcc@{}}
\toprule
\textbf{Manifestation} & \textbf{English} & \textbf{Swahili} \\
\midrule
Stereotype & 0.5128 & \textbf{0.7374} \\
Vilification & 0.6615 & \textbf{0.7208} \\
Dehumanization & 0.4453 & 0.3241 \\
Extreme Language & \textbf{0.6022} & 0.4945 \\
Lack of Empathy & 0.3724 & \textbf{0.6111} \\
Invalidation & 0.5413 & 0.5575 \\
\midrule
\textbf{Macro Average} & \textbf{0.5226} & \textbf{0.5742} \\
\bottomrule
\end{tabular}
\caption{Validation F1 per manifestation, Subtask~3.}
\label{tab:subtask3_labels}
\end{table}

English excels at political polarization (0.6682 vs.\ 0.3256), while Swahili
dominates racial/ethnic detection (0.7972 vs.\ 0.5082), reflecting corpus
composition (35.5\% racial/ethnic in Swahili vs.\ 8.7\% in English).
Gender/sexual and other categories remain hard for both languages
(F1~$<$~0.33) due to extreme rarity (2--4\% of samples), where class
weighting alone is insufficient.

Dehumanization is universally challenging (0.4453 EN, 0.3241 SW), requiring
nuanced semantic understanding to distinguish metaphorical from explicit
dehumanization.

\subsection{Ablation Studies}
\label{sec:ablation}

Table~\ref{tab:ablation_weighting} shows the impact of class weighting on
Subtask~2 English. Without weighting the model under-predicts minority labels
(41.23\% macro~F1). Class weighting improves macro~F1 by +5.98 points and
micro~F1 by +11.52 points; rare labels (religious, gender/sexual, other)
gain 10--15 F1 points while frequent labels remain stable.

We conducted the primary ablation on English for conciseness. Qualitatively,
we observed similar trends for Swahili: class weighting consistently reduced
false negatives on rare labels, consistent with prior work on imbalanced
classification \cite{johnson2019survey}.

\begin{table}[ht]
\centering
\small
\begin{tabular}{@{}lcc@{}}
\toprule
\textbf{Configuration} & \textbf{Val F1 Macro} & \textbf{Val F1 Micro} \\
\midrule
No weighting & 0.4123 & 0.3894 \\
+ Class weights & \textbf{0.4721} & \textbf{0.5046} \\
\midrule
Improvement & +0.0598 & +0.1152 \\
\bottomrule
\end{tabular}
\caption{Class weighting ablation, Subtask~2 English.}
\label{tab:ablation_weighting}
\end{table}

Table~\ref{tab:ablation_models} compares model architectures for English
Subtask~1. RoBERTa-base outperforms BERT by 2.58\%, justifying our model
choice. XLM-RoBERTa underperforms despite larger size, confirming that
multilingual pre-training dilutes English-specific representations.

\begin{table}[ht]
\centering
\small
\begin{tabular}{@{}lcccc@{}}
\toprule
\textbf{Model} & \textbf{Params} & \textbf{Val F1} & \textbf{$\Delta \%$} \\
\midrule
DistilBERT & 66M & 0.7840 & -4.82 \\
BERT-base & 110M & 0.8064 & -2.58 \\
XLM-RoBERTa & 270M & 0.8201 & -1.21 \\
\textbf{RoBERTa-base} & \textbf{125M} & \textbf{0.8322} & \textbf{--} \\
\bottomrule
\end{tabular}
\caption{Model comparison, English Subtask~1. $\Delta$ = gap from best.}
\label{tab:ablation_models}
\end{table}

\subsection{Error Analysis}

We analyzed 100 randomly sampled errors from each subtask to identify failure patterns. 

\paragraph{Subtask~1:} False positives arise from neutral political headlines
misclassified as polarizing (e.g., ``Senate passes infrastructure bill with
bipartisan support'') and hyperbolic sports comments (e.g., ``Lakers will
destroy the Celtics''). The model conflates strong language with polarization.

\paragraph{Subtask~2:} Multi-label ambiguity is the primary failure mode
(e.g., racial undertones in political statements are missed). The gender/sexual
label (2.2\% frequency) is rarely predicted despite class weighting.

\paragraph{Subtask~3:} Dehumanization is the hardest manifestation (31\% of
errors). Lack-of-empathy detection requires theory-of-mind reasoning beyond
standard transformer capabilities.

\section{Conclusion}

We presented a transformer-based system combining language-specific model
selection, class-weighted loss, and per-label threshold tuning for multilingual
polarization detection. Our system outperforms both POLAR baseline and mBERT
baselines across all Subtasks and languages, ranking 2nd for Swahili on
Subtask~3. Key error modes—political topic conflation, multi-label ambiguity,
and pragmatic inference gaps—point to future work on pragmatic reasoning,
cultural modeling, and cross-lingual transfer.

Future directions include incorporating external knowledge bases to recognize sarcasm in language and multi-task learning across subtasks to leverage complementary signals. Additionally, exploring cross-lingual transfer learning could improve low-resource language performance by leveraging knowledge from high-resource languages.

\section*{Ethical Considerations}

Polarization detection systems may pose risks if misused: classifiers could
unfairly target marginalised communities or suppress legitimate political
expression. All data were gathered from publicly accessible platforms for
research purposes. We advocate for human oversight in any real-world
deployment.

\section*{Acknowledgements}
We acknowledge and thank the organizers of Semeval-2026 Task 9 for providing dataset and evaluation framework.

\bibliography{custom}
\newpage
\appendix

\section{Hyperparameters}
\label{app:hyperparameters}

Table~\ref{tab:hyperparams} summarizes all hyperparameters used in our experiments.

\begin{table}[ht]
\centering
\small
\begin{tabular}{@{}lc@{}}
\toprule
\textbf{Hyperparameter} & \textbf{Value} \\
\midrule
\multicolumn{2}{c}{\textit{Model Architecture}} \\
Max sequence length (ST1) & 128 \\
Max sequence length (ST2/3) & 256 \\
Dropout rate & 0.1 \\
\midrule
\multicolumn{2}{c}{\textit{Training}} \\
Optimizer & AdamW \\
Learning rate & 2e-5 \\
Batch size (ST1) & 16 \\
Batch size (ST2/3) & 8 \\
Gradient accumulation & 2 (ST2/3 only) \\
Epochs (ST1) & 5 \\
Epochs (ST2/3) & 8 \\
Weight decay & 0.01 \\
Warmup ratio (ST1) & 0.10 \\
Warmup ratio (ST2/3) & 0.15 \\
Early stopping patience & 2 \\
Mixed precision & FP16 \\
\midrule
\multicolumn{2}{c}{\textit{Threshold Tuning}} \\
Search range & [0.1, 0.9] \\
Step size & 0.05 \\
\bottomrule
\end{tabular}
\caption{Complete hyperparameter configuration. ST = Subtask.}
\label{tab:hyperparams}
\end{table}

\section{Class Weight Calculations}
\label{app:class_weights}

Table~\ref{tab:class_weights_all} shows complete class weight calculations for all subtasks.

\begin{table}[ht]
\centering
\small
\begin{tabular}{@{}llcc@{}}
\toprule
\textbf{Subtask} & \textbf{Label} & \textbf{Eng} & \textbf{Swa} \\
\midrule
\multirow{2}{*}{ST1} 
& Non-polarized & 0.79 & 1.00 \\
& Polarized & 1.37 & 1.00 \\
\midrule
\multirow{5}{*}{ST2} 
& Political & 1.73 & 36.28 \\
& Racial/Ethnic & 10.45 & 1.83 \\
& Religious & 27.63 & 27.10 \\
& Gender/Sexual & 41.95 & 48.05 \\
& Other & 25.03 & 11.26 \\
\midrule
\multirow{6}{*}{ST3} 
& Stereotype & 5.54 & 1.53 \\
& Vilification & 2.75 & 1.42 \\
& Dehumanization & 7.29 & 6.69 \\
& Extreme Language & 3.18 & 3.08 \\
& Lack of Empathy & 8.11 & 2.33 \\
& Invalidation & 4.64 & 3.29 \\
\bottomrule
\end{tabular}
\caption{Complete class weights for all subtasks and languages. Weights computed as $n_{neg}/n_{pos}$.}
\label{tab:class_weights_all}
\end{table}

\end{document}